\documentclass[a4paper]{article}

\usepackage{INTERSPEECH2015}

\usepackage{graphicx}
\usepackage{amssymb,amsmath,bm}
\usepackage{textcomp}
\usepackage{mathtools}
\usepackage[english]{babel}
\usepackage{url}

\ninept

\title{Fantastic 4 system for NIST 2015 Language Recognition Evaluation}

\makeatletter
\def\name#1{\gdef\@name{#1\\}}
\makeatother \name{{\em Kong Aik Lee$^1$, Ville Hautam\"aki$^2$, Anthony Larcher$^3$, Wei Rao$^4$}\\
	{\em Hanwu Sun$^1$, Trung Hieu Nguyen$^1$, Guangsen Wang$^1$, Aleksandr Sizov$^{1,2}$}\\
	{\em Ivan Kukanov$^2$, Amir Poorjam$^2$, Trung Ngo Trong$^2$}\\
	{\em Xiong Xiao$^4$, Cheng-Lin Xu$^4$, Hai-Hua Xu$^4$}\\
	{\em Bin Ma$^1$, Haizhou Li$^1$, Sylvain Meignier$^3$}}
\address{$^1$Institute for Infocomm Research, A$^\star$STAR, Singapore
	\\$^2$ University of Eastern Finland (UEF), Finland
	\\$^3$ Universite du Maine - LIUM, France
	\\$^4$ Nanyang Technological University, Singapore}

\makeatletter

\begin{document}

  \maketitle

\begin{abstract}
   This article describes the systems jointly submitted by Institute for Infocomm (I$^2$R), 
   the Laboratoire d'Informatique de l'Université du Maine (LIUM), Nanyang Technology University (NTU) 
   and the University of Eastern Finland (UEF) for 2015 NIST Language Recognition Evaluation (LRE). 
   The submitted system is a fusion of nine sub-systems based on i-vectors~\cite{DehakKDDO11} extracted 
   from different types of features. 
   Given the i-vectors, several classifiers are adopted for the language detection task
   including support vector machines (SVM)~\cite{Campbell_svmbased}, multi-class logistic regression (MCLR), 
   Probabilistic Linear Discriminant Analysis (PLDA)~\cite{ChenLMGLD14} and Deep Neural Networks (DNN). 
\end{abstract}

  \section{Introduction} \label{sec:intro}
   
	The nine sub-systems are:
	   \begin{description}
       \item [BNF-MCLR] MCLR applied to 600-dimensional i-vectors extracted from bottleneck features (BNF2)
	   \item [BNF-PLDA-500] PLDA applied on 500-dimensional i-vectors extracted from bottleneck features (BNF1)
	   \item [BNF-PLDA-600] PLDA applied to 600-dimensional i-vectors extracted from bottleneck features (BNF2)
	   \item [BNF-SVM] SVM classifier applied on 600-dimensional i-vectors extracted from the bottleneck features (BNF2)
	   \item [LSTM] End-to-end long-short term memory neural network system
	   \item [Pair-wise DNN] Pair-wise DNN post-processing i-vectors to learn a new representation
	   \item [SDC-MCLR] Traditional i-vector system based on SDC-MFCC spectral features (SDC1) with MCLR classifier
	   \item [SDC-PLDA] PLDA applied on 500-dimensional i-vectors extracted from SDC-MFCC features (SDC2)
	   \item [Tandem-SVM] GMM-SVM system based on super-vectors extracted on 77-dimensional tandem features (BNF2 + 13 MFCC) 
	   \end{description}

  \section{Front-end} \label{sec:sys1}

	\subsection{Bottleneck features (BNF1)}
	A first set of bottleneck features is trained using the \textbf{SIDEKIT} plateform\footnote{\url{http://lium.univ-lemans.fr/sidekit}} linked to Theano.
	A GMM-HMM trained using the Switchboard Kaldi receipe provides the frame alignement that is fed to a feed-forward DNN in Theano.
	The DNN input is based TRAPS parameters computed on 31 stacked 23-dimensional mean and variance normalized filter bank features.
	The DNN is randomly initialized and uses sigmoid activation functions on all 5 layers counting 2500-2500-60-1024-2500 hidden units. The output layer 
	has 1811 senones. The resulting bottleneck features are then mean and variance normalized per file after applying an energy based voice activity detection.

    \subsection{Bottleneck features (BNF2)} \label{sec:sys1_des}
	A second set of bottleneck DNN~\cite{Richardson&Reynolds&Dehak2015} was trained using the 40-dimensional filter bank features 
	with the first and second order derivatives extracted from the switchboard landline data. 
	The features were  then applied a global mean and variance normalization followed by a per utterance mean and variance normalization 
	before feeding to the DNN. Random weight initialization is used to start the DNN training. 
	The DNN input contains 21 stacked frames rending an input layer with 2520 units. 
	Seven hidden layers including one bottleneck layer were trained. 
	Each hidden layer except the bottleneck layer has 1024 hidden units and uses the rectified linear unit (ReLU) activation function. 
	The second to last hidden layer is the bottleneck layer with 64 output units and linear outputs are extracted as the bottleneck features. 
	The output layer has 6111 units corresponding to 6111 senones obtained from the baseline speaker-independent 
	GMM-HMM system trained with 39-dimensional MFCC features (13 static features plus first and second order derivatives) 
	extracted from the switchboard landline data using Kaldi.

	\subsection{SDC-MFCC (SDC1)} \label{sec:sdc_mfcc_baseline}
	For each utterance in the dataset, 7 MFCC features and 49 Shiffted Delta Cepsta (SDC) features have been extracted. 
	To extract SDC-MFCC features, a 20 millisecond hamming window which shifted for 10 millisecond was used. 
	The SDC parameters (N-d-P-k) were configured as 7-1-3-7. These two sets of features have been concatenated to form a 
	56 dimensional SDC-MFCC feature. Standard energy VAD was applied to remove silent frames, SDC frame was decided to be speech if 70\% of the context was deemed speech by VAD.  Then, CMVN and feature warping 
	were applied over the features to remove the linear channel effects, mitigate the effect of linear channel mismatch and warp 
	the distribution of the features to the standard normal distribution.
	
	\subsection{SDC-MFCC (SDC2)}
	A second set of SDC-MFCC features has been extracted following a similar process.
	The main differences with SDC1 are that this second set of SDC-MFCC has been extracted using the publicly available \textbf{SIDEKIT} platform
	and that the VAD is applied based on  an estimated SNR.		

	\subsection{Stacked multilingual bottleneck features}
\begin{figure}[t]
		\centering
		\includegraphics[width=\linewidth]{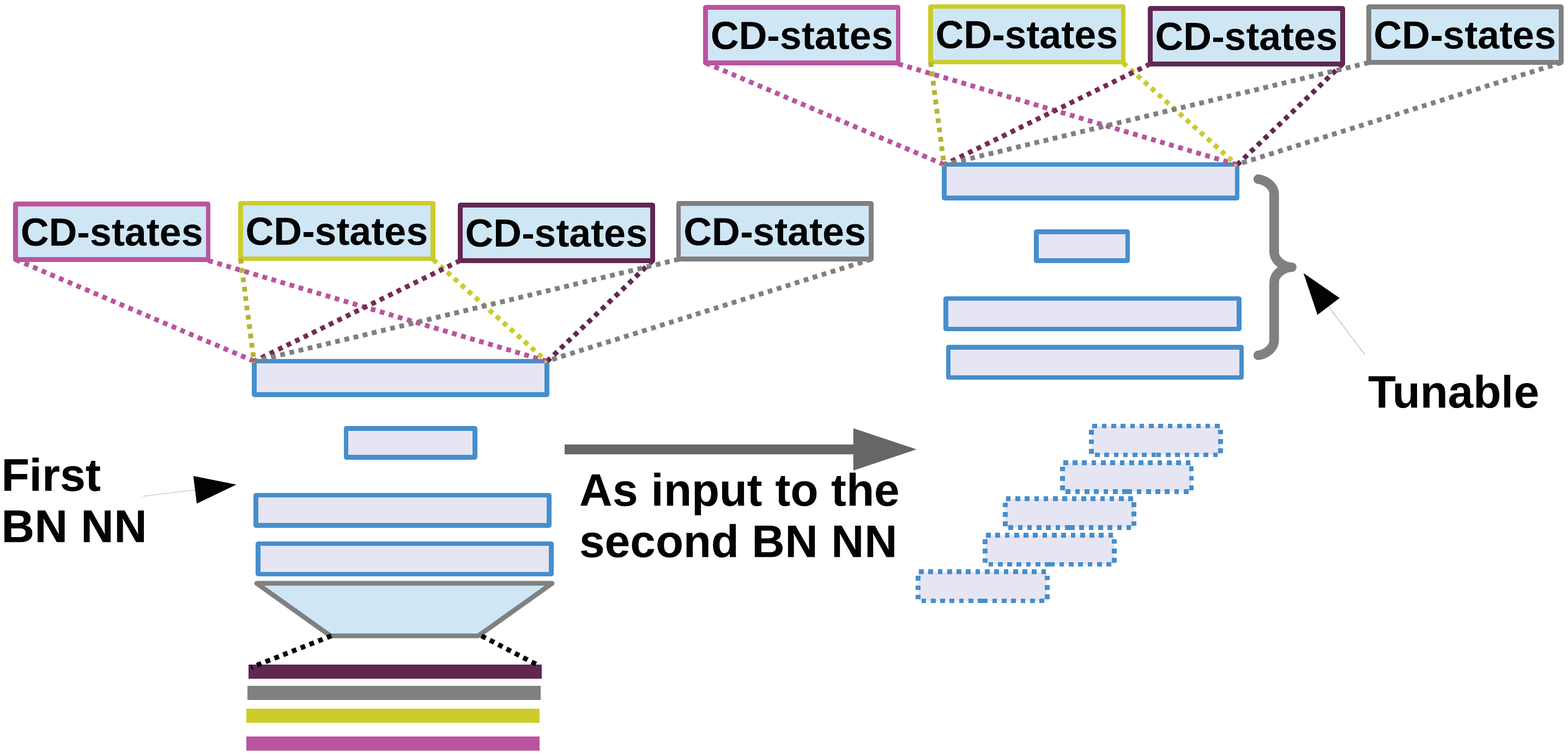}
		\caption{Illustration of the stacked multilingual bottleneck neural network training.}
		\label{fig:property_bnf}
	\end{figure}

	Prior work~\cite{Richardson&Reynolds&Dehak2015} shows that using frame-level features (BNFs) extracted from ASR Deep Neural Network (DNN) 
	with a bottleneck layer instead of shifted-delta cepstrum (SDC)~\cite{Li&Ma&Lee2015} features is very effective in NIST LRE. 
	NTU systems adopt this strategy. However, the difference with~\cite{Richardson&Reynolds&Dehak2015} is that we extracted the BNFs 
	from a stacked multilingual bottleneck neural network. Figure~\ref{fig:property_bnf} shows the framework of the stacked multilingual 
	bottleneck neural network (MBNN) training~\cite{Grezl&Karafiat2013,Grezl&Karafiat&Vesely2014,Xu&Su&Chng&Li2014,Xu&Do&Xiao&Chng2015}. 
	The class labels are changed to context-dependent (CD) states. During training, the two bottleneck (BN) neural networks (NNs) 
	are multilingually trained successively. Once the first BN NN is finished training, it is fixed as a feature transform to train the second BN NN. 
	After we finish the multilingual bottleneck network training, the two BN NNs are stacked to generate the BNFs for the unseen languages to train acoustic models, 
	realizing	MBNN based cross-lingual transfer. Original feature for training is 25-dim: 22 Mel filter-bank log energies plus 3 kaldi pitch features. 
	Hamming window and DCT transformation are applied on 11 contextual features. The first BN NN is configured as 275-1500-1500-80(BN)-1500-7887 and 
	the second 400-1500-1500-30(BN)-1500-7887. All BN layers employ linear-neurons while the other hidden-layers are sigmoid  neurons. 
	Networks are trained on GPU using gradient descent cross-entropy criterion without pre-training. 
	Note that we used monophone state labels as SBN output. To obtain better alignment triphone system was used to conduct forced 
	alignment instead, and then phone conversion is performed.
	
	Multilingual dataset released for \textit{OpenKWS}\footnote{http://www.nist.gov/itl/iad/mig/openkws.cfm} by IARPA Babel	
	program was used for pre-training stacked multilingual bottleneck neural network. The dataset was composed by 141.3 hours
	Cantonese, 78.4 hours Pashto, 77.2 hours Turkish, 84.5 hours Tagalog, 70 hours Tamil, and 87 hours Vietnamese. 
	Then, 318-hours switchboard landline data was applied to tuning the second multilingual	bottleneck neural network. 
	Only the training data provided by NIST 2015 LRE was used for estimating UBM, i-vector extractor, and language	models. Systems based on these features
were only used in extended data condition submission. 
	
	\section{i-vector extraction}

	\subsection{I2R}
	The 64-dimensional bottleneck features (BNF2) are used for extracting the i-vectors. 
	An energy-based voice activity detection (VAD) technique was applied to the raw bottleneck features to exclude the silence frames. 
	The voiced frames were then used to train a universal background model with 1024 Gaussians with diagonal covariances. 
	The diagonal UBM was then used as an initial point to train a full-covariance UBM with 1024 Gaussians. 
	The full-covariance UBM is then used to train the total variability matrix and extract the i-vectors.

	\subsection{LIUM}
	Two sets of i-vectors are extracted using \textbf{SIDEKIT} based on BNF1 and SDC2 features.
	Both i-vector extractors are based on diagonal covariance UBMs and use minimum-divergence criteria
	at each iteration. The number of distribution in the UBM is 512V for the bottleneck features and 
	1024 for the SDC features while the rank of the total variability matrix is 500 for both systems.
	i-vectors are then normalized using one itertion of the EFR algorithm \cite{Bousquet11_a}.

	\subsection{NTU - Stacked Bottleneck i-vectors}
	The i-vector extractor is trained using the 30-dim bottleneck features (BNFs) plus their 1st- and 2nd-derivatives 
	extracted from each utterance, leading to 90-dim feature vectors. It is followed by cepstral mean normalization with 
	a window size of 3 seconds. An energy based voice activity detection (VAD) method was used to remove the silence frames. 
	Then, these features are fed to the standard i-vector framework. The i-vector framework is based on a 2048 Gaussian UBM 
	with full covariance. The same data set was used to train a total variability matrix (i-vector extractor) 
	with 400 total factors. Given the i-vectors, a Gaussian back-end was used to obtain  the scores for each language.
	
	\subsection{UEF} 
	We fitted a GMM to acoustic features extracted from each speech signal. 
	Since accurately fitting a separate GMM with high number of component for short utterances is difficult, 
	parametric utterance adaptation techniques should be applied to adapt a universal background model (UBM). 
	In this approach, the i-vector framework has been employed to adapt UBM means. 
	So, a GMM with 512 mixture components was considered for the UBM. The parameters of the UBM and a 400 dimensional total variability subspace 
	were trained using  $^2/_3$ of available training data. The inferred i-vectors were whitened and length-normalized.

  \section{Core System Descriptions} \label{sec:sys2}

    \subsection{BNF-MCLR} \label{sec:mclr}
    The multiclass logistic regression (MCLR) system is based on the multi-class cross-entropy discriminative training in 
    the score vector space. To this end, i-vectors were transformed into log-likelihood score vectors through a set of 
    Gaussian distributions, each representing the distribution of the language class in the i-vector space. 
    As the amount of data is extremely imbalance among classes, with some languages limited to less than an hour of speech, 
    we trained a global covariance matrix where language-specific covariance could be derived with a smoothing factor of 0.1. 
    Given a test i-vector, a score vector is obtained by concatenating the log-likelihood scores from these Gaussian distribution.
    Discriminative training is further applied on the score vector. The multiclass FoCal toolkit was used for this purpose.

    \subsection{SDC-MCLR} 
    A linear Gaussian back-end was trained using $^2/_3$ of available training data to compute log-likelihood scores for each utterance. 
    To calibrate the scores, the scores of evaluation set were applied to a linear logistic regression which was trained with scores 
    of $^1/_3$ of training data. Finally, calibrated scores were scaled to log-likelihood ratio. 
    In these experiments, FoCal toolkit has been used to calibrate the scores.

	\subsection{BNF-PLDA-500}
	i-vectors extracted by the LIUM have been used to train a Gaussian-PLDA of rank 70. The resulting scores
	computed for each language are fed to a linear Gaussian Backend. Resulting scores are then rescaled for each of the cluster.
	Note that this system is entirely build on the \textbf{SIDEKIT} 
platform and that a tutorial will be released post eval to 
	provide the entire training process.
	
    \subsection{BNF-PLDA-600} \label{sec:sys2_des}
	The bottleneck i-vectors are used in this system.The i-vectors extracted from the files shorter than 1 second 
	were removed from the training. The i-vectors are the applied with whitening and length norm. For each language cluster,  
	a domain-adapted Simplified PLDA~\cite{Garcia-RomeroM14} (weighted likelihood approach) with 20-dimensional latent subspace is trained.
	The training data for each language within a cluster is used  as the enrollment data and a adapted-prior (single Gaussian) 
	for PLDA~\cite{ChenLMGLD14} is computed. The  Gaussians are then fine-tuned with the MMI procedure~\cite{Alan14}.

	\subsection{BNF-SVM} \label{sec:svm-i-vector}
	We also use i-vector features to construct two more SVM systems. 
	The i-vectors are directly feed into SVM system and train individual language model.  
	The pair-wise testing method is used to generate the individual evaluation segment scores.

	\subsection{End-to-end Long-short term memory (LSTM)} \label{sec:end_to_end_lstm}
	The audio data was preprocessed into 20 ms frames, overlapped by 10ms, into 13 Mel-frequency Cepstrum Coefficients (MFCCs) 
	from 24 filter-banks using \textbf{SIDEKIT}. The first and second derivatives of all coefficients was also concatenated to the original 
	coeffecients, giving a 39-dimensional vector. The coefficients were normalized to have mean 0 and standard deviation 1 over 
	each utterances. Additionally, LSTM need input is a sequence, we roll MFCC into a sequence of 20 frames. 
	This process is performed for each speech file, and the remain frames are truncated.
	We used bidirectional recurrent neural network \cite{birnn}, in combination with LSTM \cite{lstm}, 
	to bridge long time lags with access to the past and future context of the signal. We chose bidirectional LSTM (BLSTM) 
	because our experiments with unidirectional LSTM gave worse results on the task. The network structure is specified in following table:
	\begin{center}
		\begin{tabular}{ | l | l | l |}
			\hline
			\textbf{Type} & \textbf{\# Hidden Units} & \textbf{Note} \\
			\hline
			Input layer   &  & (batch size, 20, 39)\\
			\hline
			BLSTM         & 250 &  \\
			\hline
			BLSTM         & 250 &  \\
			\hline
			BLSTM         & 250 & output is delayed by 15\\
			\hline
			Projection & 512 & Activation: rectifier \\
			\hline
			Output          & 20  & Softmax for 20 languages \\
			\hline
		\end{tabular}
	\end{center}
	All the network's parameters were initialized using Glorot's uniform mentioned in \cite{glorot}. 
	BLSTM layers used sigmoid as inner activation and tanh for output activation.
	For training the network we used categorical crossentropy objectives (i.e. one-vs-all criterion) 
	and RSMprop optimizer \cite{rmsprop}, which utilizes the magnitude of recent gradients to normalize the 
	gradients for each backward training step. In order to prevent overfitting training set, we used 4 different techniques:
	\begin{itemize}
		\item Early stopping based on generalization loss of tuning set \cite{earlystop}.
		\item L2 regularization (weights decay = 1e-4) on 2 Fully connected layers
		\item Dropout with probability of ignoring an activation is uniform(0.3) \cite{dropout}
		\item Adding random Gaussian noise to perturb the weights.
	\end{itemize}
	Our experiments showed that training deep model (i.e. BLSTM) on biased dataset will cause the whole network 
	converges to the classes with higher proportion of samples. Hence, the training process was carefully executed with a fixed scenarios:
	\begin{enumerate}
		\item Training 2 epochs with RSMprop, learning rate = 0.001, without noise and dropout, only L2 regularization and early stop enable.
		\item The next 2 epochs with almost the same configuration except dropout probability = 0.3 and learning rate = 0.0001.
		\item Learning rate reduced to 0.00001, with same configuration, 10 more epochs (early stop will stop the algorithm in 
		advance before any strong overfit, and only the best weights (i.e. gave the best result on validation set) was saved.
		\item After that, we used 0.075 Gaussian noise to perturb the network for 10 more epochs without any dropout.
		\item Then, we performed cyclic training to reduce the effect unbalanced training set. 
		The dataset was organized into 2 set, first set with full of training data, the second one was undersampling from training 
		set with balanced distribution of all languages. The training configuration was as follow: every 2 epoch of training with full set, 
		do 2 other epoch with undersampled set (learning rate = 0.00001, with dropout and noise used in turn). 
		This procedure was repeated until no more improvement achieved and cancelled by early stop.
	\end{enumerate}
	Taken into account the biased distribution of training set, the network under training process was strongly influenced by the 
	languages with large amount of data. We used prior knowledge from training distribution to calibrate the softmax score before 
	calculated likelihood ratio. We only did calibration on the 3 clusters with the worst performance, addressed the fact that 
	the network was already biased to provide the best result for 3 other clusters.

	\subsection{Pair-wise DNN} \label{sec:pwdnn}
\begin{figure}[t]
		\centering
		\includegraphics[width=\linewidth]{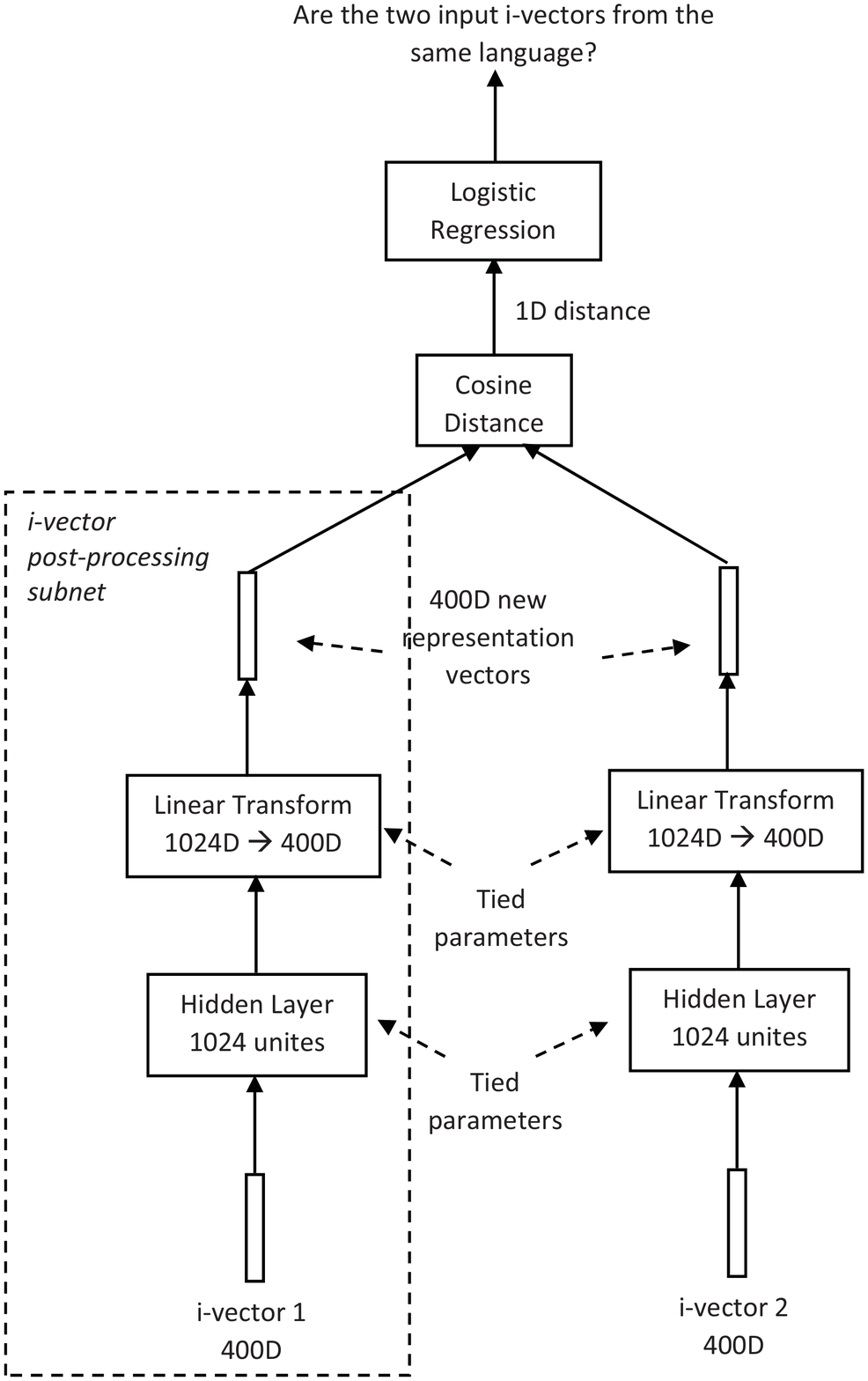}
		\caption{The diagram of the DNN post-processing on i-vectors.}
		\label{fig:dnn_sys}
	\end{figure}

	Instead of using Gaussian back-ends, DNN post-processing is adopted to get a new representation from the i-vectors. 
	The diagram of the DNN post-processing is shown in Figure~\ref{fig:dnn_sys}. Each training sample consists of two 
	input i-vectors and one label. The label is 1 if these two i-vectors are from the same language and 0 otherwise. 
	Two i-vectors will be processed by the i-vector post-processing subnet which consists of one or more hidden layers 
	and one linear transform. During training the gradient of the left and right subnets are computed individually, 
	but their parameters tied and updated together to make sure that the two subnets have the same parameter at all time. 
	The output of the subnets are new representation vector of the respective sentences and will be used for language recognition. 
	The subnets are trained such that simple cosine distance is able to tell whether the two input i-vectors are from the same language. 
	The network is trained layer-by-layer. For example, the first hidden layer is firstly trained  until convergence. Then, we add the second hidden 
	layer or the linear transform layer and train the whole network until convergence. The training usually converges after 10-20 epochs.
	
	To create the training samples for the DNN, both positive i-vectors pairs where the two i-vectors are from the same languages 
	and negative pairs are randomly generated. In the most basic setting, for each i-vector, we create one positive pair and one negative pair. 
	To create more balanced training samples, we generate more positive/negative pairs for languages with little i-vectors and this 
	is found to be helpful on the tune set. We run the pair generation algorithm 20 times to create about 3 million training pairs.

    \subsection{Tandem-SVM} \label{sec:svmubm}
	The system uses the bottleneck feature super-vectors to construct kernels of support vector machines (SVMs). 
	Given a language speech data, a GMM is estimated by using MAP adaptation of the means of the UBM. 
	The global 1024 Gaussian mixture component model is trained based on the NIST's provided developed dataset. 
	The means of mixture components in the GMM are concatenated to a GMM supervector. 
	The features used is the 64 dimension bottleneck feature with additional 13 mfcc,
	which result in a feature space expansion from 64 to 56636, and 77 to 78848 in dimension, respectively.

	\section{Extended data condition}
Compared to the core systems, the contrastive systems rely on the stacked bottleneck feature i-vectors including MCLR 
	in section~\ref{sec:mclr}, SVM-Ivector as in section~\ref{sec:svm-i-vector}, SVM-UBM-MFCCBN77 in section~\ref{sec:svmubm} 
	by replacing MFRCCBN77 with the 30-dimensional stacked bottleneck features and the pair-wise DNN in section~\ref{sec:pwdnn}.

\section{Fusion and submission}
Fusion was decided to be the typical multi-class logistic regression. 
The development set of scores was divided into two parts, where 
model parameters were estimated with one part and applied to the 
held-out part. Regularization was experimented with, but as 
the improvement on held-out set was negligible it was discarded and 
final fusion scores were estimated without regularization. Fusion was
applied to evaluation set and scores were turned into log-likelihood ratios
on per language cluster basis.

  \newpage
  \eightpt
  \bibliographystyle{IEEEtran}

  \bibliography{mybib}

\end{document}